\documentclass[acmsmall]{acmart}
\usepackage[capitalise]{cleveref} 
\usepackage{multirow} 
\AtBeginDocument{%
  }

\setcopyright{acmlicensed}
\copyrightyear{2018}
\acmYear{2018}
\acmDOI{XXXXXXX.XXXXXXX}

\acmJournal{JACM}
\acmVolume{37}
\acmNumber{4}
\acmArticle{111}
\acmMonth{8}

\begin{document}

\title{Towards certifiable AI in aviation: landscape, challenges, and opportunities}
\thanks{The European Union partially supported the work reported in this paper in the project SustainML under grant agreement number 101070408.}
\author{Hymalai Bello}
\email{hymalai.bello@dfki.de}
\orcid{--}
\affiliation{%
  \institution{German Research Center for Artificial Intelligence}
  \streetaddress{Trippstadter 122}
  \city{Kaiserlautern}
  \state{Rhineland-Palatinate}
  \country{Germany}
  \postcode{67663}
}
\author{Daniel Geißler}
\email{Daniel.Geissler@dfki.de}
\orcid{--}
\affiliation{%
  \institution{German Research Center for Artificial Intelligence}
  \streetaddress{Trippstadter 122}
  \city{Kaiserlautern}
  \state{Rhineland-Palatinate}
  \country{Germany}
  \postcode{67663}
}
\author{Lala Ray}
\email{Lala_Shakti_Swarup.Ray@dfki.de}
\orcid{--}
\affiliation{%
  \institution{German Research Center for Artificial Intelligence}
  \streetaddress{Trippstadter 122}
  \city{Kaiserlautern}
  \state{Rhineland-Palatinate}
  \country{Germany}
  \postcode{67663}
}

\author{Stefan Müller-Divéky}
\email{stefan.mueller-diveky@diehl.com}
\orcid{--}
\affiliation{%
  \institution{Diehl Aerospace GmbH}
  \streetaddress{An d. Sandelmühle 13}
  \city{Frankfurt}
  \state{Hessen}
  \country{Germany}
  \postcode{60439}
}
\author{Peter Müller}
\email{peter.mueller@diehl.com}
\orcid{--}
\affiliation{%
  \institution{Diehl Aerospace GmbH}
  \streetaddress{Alte Nußdorfer 23}
  \city{Überlingen}
  \state{Baden-Württemberg}
  \country{Germany}
  \postcode{88662}
}
\author{Shannon Kittrell}
\email{Shannon.Kittrell@dfki.de}
\orcid{--}
\affiliation{%
  \institution{German Research Center for Artificial Intelligence}
  \streetaddress{Trippstadter 122}
  \city{Kaiserlautern}
  \state{Rhineland-Palatinate}
  \country{Germany}
  \postcode{67663}
}
\author{Mengxi Liu}
\email{Mengxi.Liu@dfki.de}
\orcid{--}
\affiliation{%
  \institution{German Research Center for Artificial Intelligence}
  \streetaddress{Trippstadter 122}
  \city{Kaiserlautern}
  \state{Rhineland-Palatinate}
  \country{Germany}
  \postcode{67663}
}
\author{Bo Zhou}
\email{Bo.Zhou@dfki.de}
\orcid{--}
\affiliation{%
  \institution{German Research Center for Artificial Intelligence}
  \streetaddress{Trippstadter 122}
  \city{Kaiserlautern}
  \state{Rhineland-Palatinate}
  \country{Germany}
  \postcode{67663}
}
\author{Paul Lukowicz}
\email{Paul.Lukowicz@dfki.de}
\orcid{--}
\affiliation{%
  \institution{German Research Center for Artificial Intelligence}
  \streetaddress{Trippstadter 122}
  \city{Kaiserlautern}
  \state{Rhineland-Palatinate}
  \country{Germany}
  \postcode{67663}
}

\renewcommand{\shortauthors}{Bello et al.}

\begin{abstract}
Artificial Intelligence (AI) methods are powerful tools for various domains, including critical fields such as avionics, where certification is required to achieve and maintain an acceptable level of safety.
General solutions for safety-critical systems must address three main questions: Is it suitable? What drives the system's decisions? Is it robust to errors/attacks?
This is more complex in AI than in traditional methods.
In this context, this paper presents a comprehensive mind map of formal AI certification in avionics. 
It highlights the challenges of certifying AI development with an example to emphasize the need for qualification beyond performance metrics. 
\end{abstract}



\received{20 February 2007}
\received[revised]{12 March 2009}
\received[accepted]{5 June 2009}

\maketitle

\section{Introduction}
Artificial intelligence (AI) is revolutionizing the avionics field (AI in aviation), offering many advantages and challenges. 
This fusion can increase efficiency, enhance safety, and improve passenger experience.  
AI in aviation currently focuses on AI-for-Cabin and non-critical tasks.
On the other hand, AI-for-non-Cabin tasks encompass artificial intelligence techniques for the operation of the aircraft, for example, vehicle management or flight control/guidance/management system functions.
AI-for-non-Cabin tasks are therefore subject to stringent certification requirements and a thorough and explainable understanding of the target tasks and AI methods to ensure the safety of passengers, flight crew, and aircraft.
Moreover, the scope of AI-for-non-Cabin tasks ranges from communication, radar, digital electronics, integrated avionics systems, and navigation, to advanced traffic detection systems, all being considered critical tasks. 

To develop any application in the safety-critical aviation sector, certain standards must be followed to meet the industry's safety and security restrictions. 
The authorities recognize several industry standards as acceptable means of compliance. 
For example, for system-related aspects, the Guidelines for Development of Civil Aircraft and Systems (ARP4754B) are available \cite{ARP4754B}. 
For software aspects, the Software Considerations in Airborne Systems and Equipment Certification (DO-178C) exists \cite{DO-178C}. 
In the case of data certification, there are also the Standards for Processing Aeronautical Data (DO-200B) \cite{DO-200B}.
The main limitation of these guidelines is that they do not entirely cover the challenges of AI-enabled systems. 
This led to the European Union Aviation Safety Agency (EASA) to work on defining equivalent methods for the safe use of machine learning (ML) approaches.
In 2024, the EASA published the Artificial Intelligence Concept Paper: Guidance for Level 1 \& 2 machine learning applications \cite{EASAConcept} in response to the EU AI Act Chapter III \cite{EUAIACT}.
It defines four AI certification building blocks, following the Ethics Guidelines for Trustworthy AI \cite{ai2019high}: 
\begin{itemize}
    \item AI Trustworthiness Analysis
    \item AI Assurance
    \item Human Factors for AI
    \item AI safety risk mitigation
\end{itemize}
Furthermore, the paper focuses on Level 1 AI (assistance to humans) and Level 2 AI (human-AI teaming), covering the scope of the Rule Making Task RMT.0742 to be executed at the end of 2027. 
The guideline for Level 3 AI (advanced automation) is estimated to be ready at the end of 2025. 
Additionally, EASA, in cooperation with industry partners, has published its final report of "Machine Learning Application Approval" (MLEAP)\cite{MLEAP}. 
These documents present basic guidance standards for the aviation industry's certification of AI methods in Europe.
This is accompanied by the recently released "Roadmap for Artificial Intelligence Safety Assurance" by the Federal Aviation Administration (FAA) of the United States\cite{FAARoadmap}, in compliance with Executive Order 14110: Safe, Secure, and Trustworthy Development and Use of Artificial Intelligence\cite{USAPresident}.
The guideline uses common methods such as configuration management and validation.
These classical methods are complemented by new techniques that address the specific characteristics of deep learning (DL) systems, including data collection through the training phase to DL deployment. 
This should be seen as a complement to established development methods and standards.  

As stated previously, the EASA provides a basic certification guideline for Level 1 \& Level 2 AI.
It raises questions about how to translate aviation requirements to specific areas of AI research.
AI research is a broad and multidisciplinary area, and the same question often has many answers. 
Moreover, the complexity of avionics combined with recent massive AI methods (at parameter and complexity levels) leads to a very beneficial and risky fusion.
This makes concise and correct cooperation between industry and academia crucial. 
This is a bidirectional communication channel. 
Hence, researchers need to know what questions to ask and at what stage of AI development to ask them.
On the other hand, the industry urgently needs to comprehend how next-generation AI approaches fit into the certification cycle. 
And, jointly, researchers and industry need to identify current AI approaches that fall outside the scope of certification, which means they urgently require strategies to make them reliable.
In this context, this work provides a comprehensive introduction to the roadmap toward AI certification in avionics, following the certification structure proposed by EASA.
The idea is to understand the requirements of avionics in AI terminology and vice versa to reveal the current status of AI certification in avionics, highlighting the limitations of current methods.
To achieve these objectives, the certification roadmap is reduced to its main components at each step, along with the most advanced AI methods to address them. 
Furthermore, the challenges of the classical AI development cycle are shown through an example of a widely used AI model. 
The work concludes with a list of limitations encountered for AI certification in avionics.

The structure of the paper has been decided to provide the research community and industry with a mental mapping of the current status of formal AI certification in aviation as follows:
\cref{chap:Background} provides an overview of the ML development cycle, and introduces the current approach to AI certification in avionics.
The following four sections are the main certification blocks defined according to the EASA concept paper \cite{EASAConcept}.
\cref{chap:TA} present the trustworthy analysis for AI systems in avionics, including \textit{ethical aspects} and \textit{safety and security risk management} objectives.  
This is enhanced by state-of-the-art (SOTA) tools that can be used to assess the safety and ethics of AI. 
\cref{chap:AIA} summarizes the AI assurance cycle in aviation with an overview of the SOTA methods based on the W-shape model from EASA.
\cref{chap:HFAI} focuses on the human factor of AI, and its relevance to the creation of an efficient human-AI collaborative and cooperative team.  
\cref{chap:AIS} introduces the need to address the mitigation of AI security risks that could arise from partially meeting the above certification blocks.
Moreover, in \cref{chap:Example} an example is used to show the challenges of classical certification of the AI research cycle.  
Next, \cref{chap:Discussion} presents a list of limitations and insights towards certifiable AI in aviation. 
Finally, \cref{chap:Conclusion} concludes the work.

\section{Background}
\label{chap:Background}
\begin{figure}[t!]
    \centering
    \includegraphics[width=\textwidth]{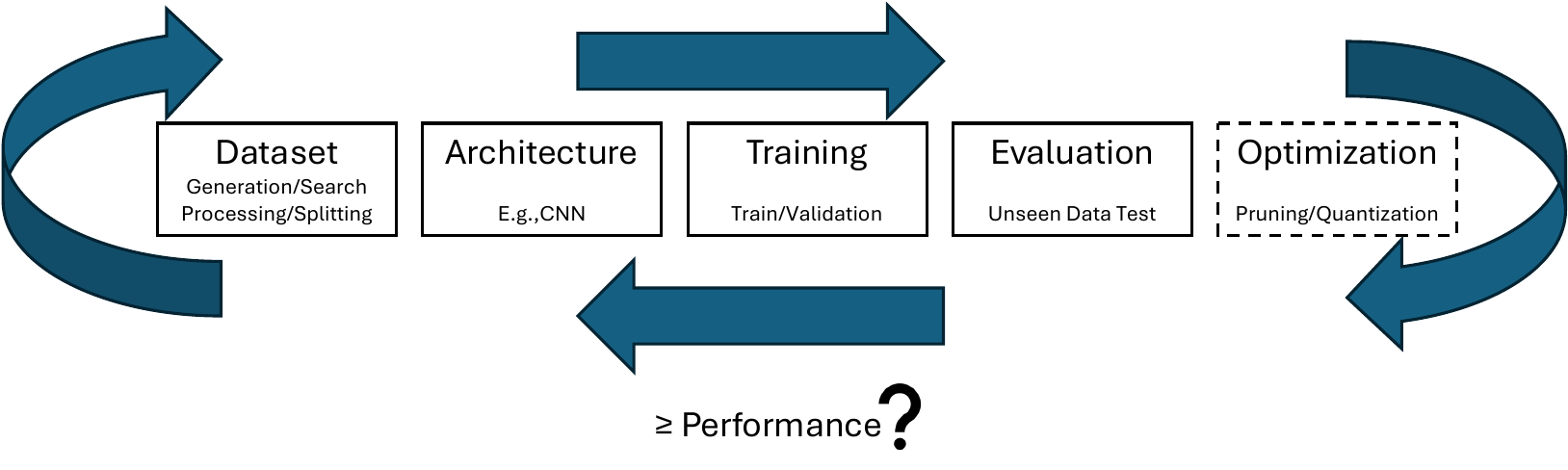}
    \caption{General pipeline for developing Deep Neural Networks (DNN)}
    \label{fig:DNNPipeline}
\end{figure}
\subsection{Machine learning and artificial neural networks}
In contrast to conventional control systems, neural networks (NN) elevate computer capabilities by facilitating learning from experience, also called data-driven methods. 
This transformative approach empowers computers to make decisions and predictions without explicit programming. 
These networks emulate the adaptability of the human brain, introducing a dynamic and intuitive dimension to computing where machines evolve and respond intelligently to diverse scenarios \cite{samek2021explaining}.
Deep Neural Network (DNN) is a type of NN with multiple layers between the input and output layers. 
These additional layers allow the network to learn complex patterns and non-linear relationships in the data to adapt to today’s complex use-case scenarios and rising amounts of data. 
This makes them incredibly versatile and powerful for a wide range of tasks\cite{larochelle2009exploring}.

\cref{fig:DNNPipeline}, depicts the general pipeline for developing DNN.
It is an iterative process that finishes when the required performance is reached. 
The pipeline applies to classification/regression tasks for various learning paradigms. 
The generation or search for the dataset is the first step. 
The data source's main requirement is to represent the modality and purpose of the specific use case. 
Typically, the dataset will require preprocessing. 
For example, filtering noisy data, handling missing values, and normalization. 
This helps the model to converge faster and generalize better. 
Dataset preparation includes splitting it into three partitions: 1. train; 2. validation; 3. test datasets.
The second step is the NN architecture design, for data scientists is a crucial phase, where the selection of an appropriate structure significantly influences the model's performance. 
The selection depends on the nature of the task, the data type involved, and the problem to solve. 
Usually, each architecture has specifications, which need to be considered to solve the envisioned task.
For example, it is typical to select the number and type of layers, patterns, and activation functions to ensure that the neural network learns and generalizes effectively from the data provided.
Next, is the training, where the model is trained by iteratively presenting batches of data through the network for a specified number of epochs. 
Each epoch represents a complete pass through the entire training dataset. 
During each epoch, the DNN optimizer adjusts the model's weights to minimize the chosen loss function through backpropagation. 
The optimizer calculates the gradients of the loss concerning the model's parameters, indicating how much each parameter contributed to the error. 
The optimizer then updates the model's weights in the opposite direction of the gradients, gradually improving the model's ability to make accurate predictions throughout training.
After each epoch, the validation set is passed through the network to monitor the loss for passing “unseen” data through the model to prevent overfitting (the model does not generalize but memorizes the data). 
This information can further be used to adjust the model's hyperparameters and detect convergence.
After the training is completed, the final model is evaluated on the test dataset to ensure the required model’s performance on previously “unseen” data is fulfilled.
This signals the breaking point of the loop and the proposed solution is completed.
The decision is based on performance metrics such as accuracy, recall, precision, and F1 score\cite{johnson2019survey}. 
This excludes trustworthy analysis, AI assurance, human factors, and safety and risk management of the solution.

Optimization of the model is an optional step but imperative for the efficient deployment of the solution on hardware-constrained devices\cite{menghani2023efficient}. 
It can be integrated into the training process or applied after training as a fine-tuning step. 
It includes methods such as pruning and quantization. 
Pruning removes superfluous or unnecessary connections within the neural network with reduced impact on the performance. 
Identifying and eliminating less meaningful connections makes the model lightweight while maintaining the best predictive capabilities possible.
On the other hand, DNNs are trained in floating-point 32-bit arithmetic to take advantage of the wider dynamic range.
Quantization is a technique that reduces the bit precision of the model's parameters.
The model's memory footprint is reduced by representing weights and activations with fewer bits, leading to faster inference times and reduced resource requirements during deployment.
There is also the option of using quantized parameters to train the model, which is called quantization-aware training.  
The idea is to model the effect of quantization, which allows for increased accuracy at the time of inference compared to post-quantization methods\cite{liang2021pruning}. 
The selection of the optimization strategy is part of the qualification process of the DNN model.
This impacts the operational performance of the final model.
It is important to note that at this stage the NN is frozen, and any changes will reopen the entire qualification process.
Therefore, it is imperative to ensure that the optimization method does not render the DNN useless.

Overall, the DNN development cycle does not usually include formal qualification steps, as it focuses on performance metrics and neglects ethical aspects and safety risk management. 
Moreover, the research community around the world is moving forward without knowing how to regulate it and ensure the realistic application of the methods in the future. 
This is a major drawback for leveraging the advantages of DNNs in critical domains such as aviation. 
Faced with this limitation, researchers and industry must join forces to stimulate AI certification-conscious research. 
 \begin{figure}[!t]
    \centering
    \includegraphics[width = \textwidth]{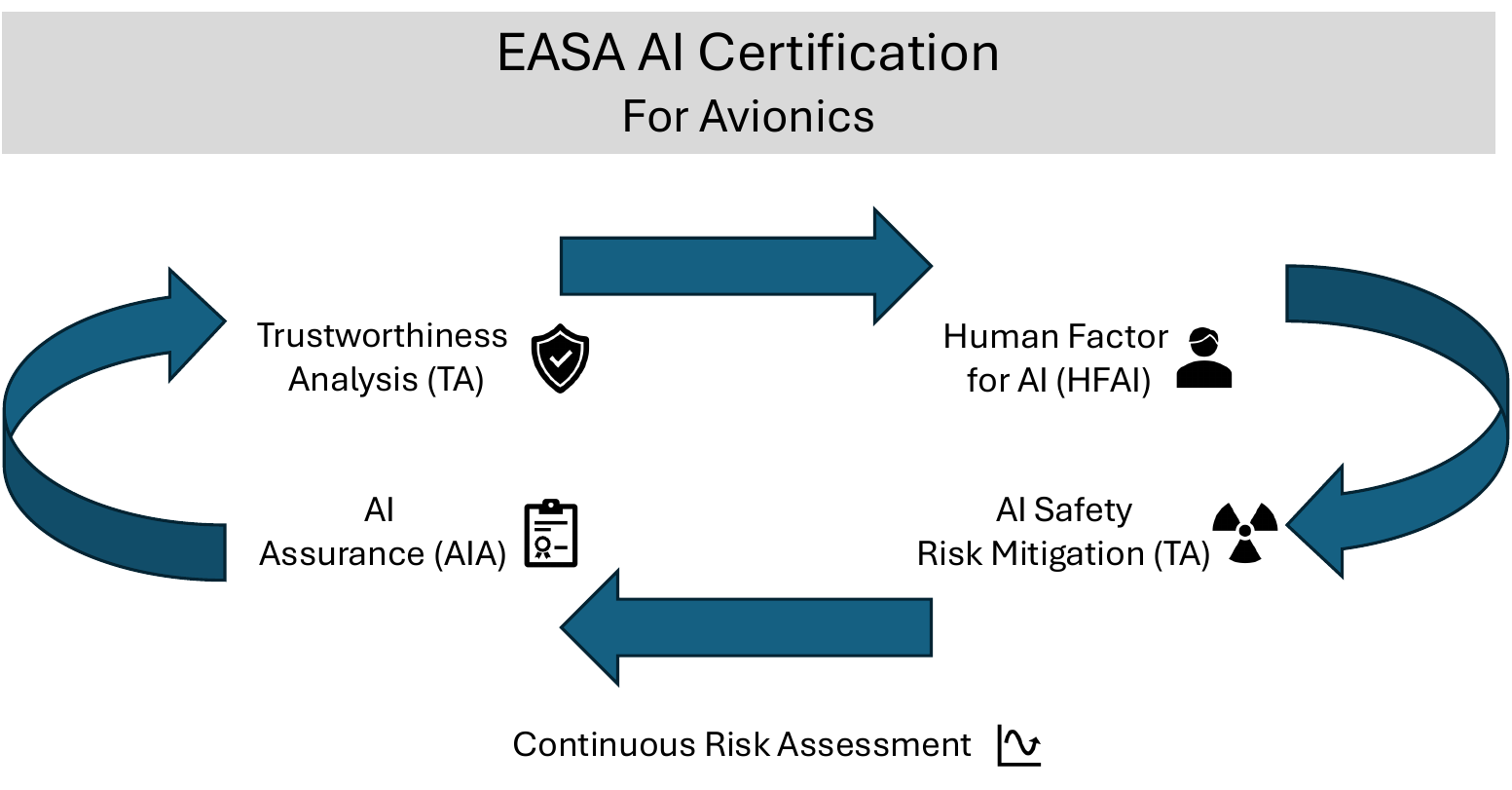}
    \caption{AI for avionics certification blocks according to the  European Union Aviation Safety Agency (EASA) Artificial Intelligence Concept Paper: Guidance for Level 1 \& 2 machine learning applications\cite{EASAConcept}. }
    \label{fig:EASACertification}
\end{figure}

\subsection{Avionics}
Avionics are the electronic systems used on an aircraft. 
It is derived from "aviation electronics", which includes communication, navigation, flight control, monitoring, display, and aircraft management systems. 
These systems continually evolve to improve efficiency, cost, safety, and risk management\cite{moir2013civil}. 
The aviation field is currently undergoing an AI revolution \cite{he2021statistical}. 
The AI can assist in predictive maintenance, for example, automatic visual inspection (AVI)\cite{yasuda2022aircraft}. 
This helps operators with faster damage detection, holistically reducing the time expended in maintenance by detecting damages in the early stage. 
Additionally, Air Traffic Control Speech-to-Text Technology (ATC-STT) aims to translate spoken instructions into text, thereby increasing safety\cite{badrinath2022automatic,lin2020unified,lin2021atcspeechnet}. 
Moreover, the Airborne Collision Avoidance System for Unmanned Aircraft (ACAS) can benefit from faster object detection and warning response times to avoid intruder aircraft in time\cite{panoutsakopoulos2022towards,choi2017two}. 
The above are examples of aviation use cases that can have a highly beneficial/risky impact when using an AI model. 

The avionic use cases involve complex systems with high dimensionality. 
To overcome the complexity, the first step in the task-solving process is to define the operational domain (OD). 
OD captures the operation conditions under which a solution/product is specifically designed to function as intended. 
The OD is defined as a set of constraints and requirements for a specific purpose (e.g., ACAS) \cite{whitworth20235g}. 
Compliance with the OD guarantees the robustness of the design. 
In the case of an AI system, the terminology is expanded to OD and the operational design domain (ODD) to include the formal requirements dealing with AI-based systems. 
The OD focuses on the entire system and the ODD focuses on the AI/ML constituent. 
AI/ML constituent includes the collection of hardware and software used to support an AI-based subsystem. 
The ODD provides a framework for the selection, collection and preparation of data during the learning phase. 
It also describes the requirements for the monitoring of data in operations. 
A precise definition of the ODD is a prerequisite for the quality, completeness, and representativeness of the datasets involved in the learning assurance process. 
A particular requirement of safety-critical avionics systems is that they must be certified.
On the other hand, DNNs are enormously complex methods, especially not very transparent or easy to interpret.
DNNs are very advantageous and risky at the same time, so combining them with avionics will pose many challenges. 
Hence, in the next section, AI certification for avionics is introduced. 

\subsection{AI certification in Avionics}

Certification of any system intended to be used in avionics is required to achieve and maintain an acceptable level of safety.
One of the prominent means of compliance includes the Software Consideration in Airborne Systems and Equipment Certification (DO-178C). 
This is the primary document used by the most famous certification authorities such as EASA for Europe, the Federal Aviation Administration (FAA) for the United States, and Transports Canada Civil Aviation (TCCA) to demonstrate design assurance for software items in avionics systems \cite{rierson2017developing}. 
For hardware certification, the Design Assurance Guidance for Airborne Electronic Hardware (DO-254) exists\cite{hilderman2007avionics,fulton2014airborne}, in addition to the Environmental and Test Procedures for Airborne Equipment (DO-160)\cite{sweeney2015understanding}, among others\footnote{https://skybrary.aero/safety-regulations/certification}. 

High certification standards are also to be expected when AI meets avionics. 
As shown in \cref{fig:DNNPipeline}, the DNN pipeline suffers from a lack of qualification. 
In this context, the EASA Concept Paper \cite{EASAConcept} intends to guide Level 1 \& Level 2 AI development in aviation. 
Level 1 relates to human assistance. 
The requirements for this level include learning assurance, AI explainability, and continuous safety and security risk assessment. 
Level 2 requires additional measures such as an ethics-based assessment and human-AI teaming. 
Furthermore, the EASA defines Level 3 AI as advanced automation and beyond. 
This upper level is the scope of the EASA's future work and the guidance for Level 3 is expected in 2025. 
It considers the extension to reinforcement and symbolic learning, statistical and hybrid AI combined with human-AI supervision, and unsupervised automation safety risk mitigation. 
It should be noted that the guidance of EASA for Level 1  \& Level 2 is still under discussion and is expected to be finalized by 2026.
And, the first expected AI approval for Level 2/3A will be in 2035, so AI certification is still in its infancy. 
For Level 1 \& Level 2, the \cref{fig:EASACertification} depicts the iterative certification flow of AI for aviation purposes. 
It presents four main blocks: Trustworthiness Analysis (TA), AI Assurance (AIA), Human Factor for AI (HFAI), and AI Safety Risk Mitigation (AIS), and based on these four blocks this work is divided. 
The following sections summarize each of the blocks individually to describe their main purposes, along with a review of the SOTA approaches related to each block is presented.

\section{Trustworthiness Analysis (TA)}
\label{chap:TA}
\begin{figure}[!t]
    \centering
    \includegraphics[width=\textwidth]{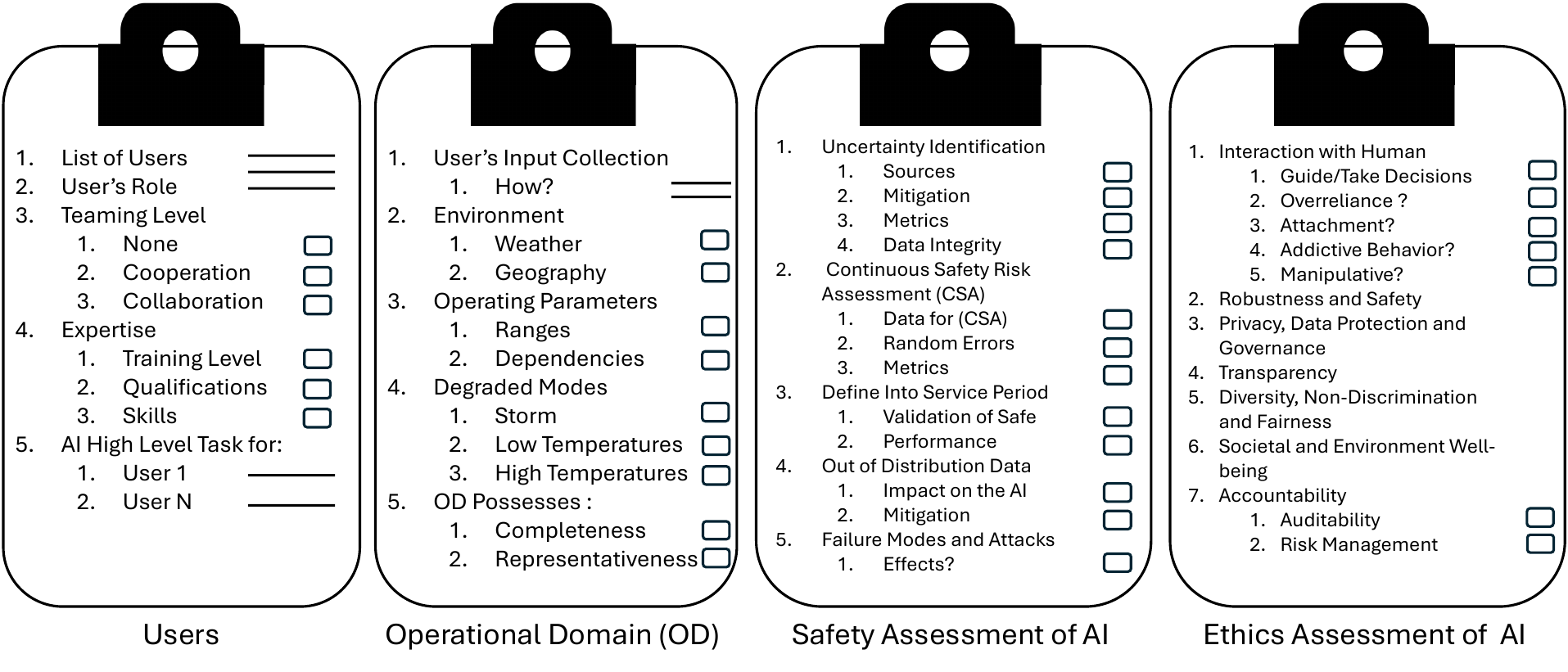}
    \caption{Objectives overview for a trustworthy analysis (TA) of artificial intelligence (AI) solutions in avionics}
    \label{fig:TAFigure}
\end{figure}
Trustworthiness Analysis (TA) for AI is independent of the type of learning algorithm; supervised, unsupervised, or reinforcement learning (RL). 
For this analysis, the system is considered as a whole, rather than considering only the separate AI subsystem. 
It comprehends two basic principles; \textit{Ethical Aspects} and \textit{Safety and Security Risk Management}. 

The first step in the TA is to determine the AI system and how it is defined. 
In aviation, the system definition depends on the specific application domain. 
The system is composed of interrelated items to perform a function at the aircraft level \cite{ARP4761A}.
For the Air Traffic Management and Air Navigation Services (ATM/ANS) domain and according to the Regulation (EU) 2017/373 \cite{RulesATMANS}, a system is defined as a combination of procedures including human resources, equipment, hardware, and software.
Therefore, certifying an AI-based solution for avionics requires a clear understanding of the scope of the system to treat it as a whole. 
Moreover, the developer needs to identify the classification of the AI application; 1. Level 1A Human augmentation; 2. Level 1B Human assistance; 3. Level 2A Human-AI cooperation; 4. Level 2B Human-AI collaboration; 5. Level 3A Advanced supervised automation; 6. Level 3B Advanced unsupervised automation. 
The first three levels (1A, 1B, and 2A) give complete authority to the user. 
At level 2B the user has partial authority. 
At level 3A the user's authority is limited upon alerting. 
At level 3B the AI has full authority. 
The correct classification is crucial for the safety and ethics assessment of the AI constituent. 

Trustworthiness in avionics includes aspects, such as precise user description, the completeness of OD and ODD description requirements, and thorough safety and ethics assessment. 
The user description, OD, and ODD requirements are currently in the hands of human experts. 
\cref{fig:TAFigure}, presents an overview of objectives to meet for a trustworthy AI development, following the EASA guidance \cite{EASAConcept}. 
It shows a collection of checklists considering users, OD, safety, and ethics assessment of AI. 
In \cref{tab:SafetyEthics} are shown state-of-the-art tools that can be used for ethical and safety assessment of AI.

\begin{table}[!t]
    \centering
    \footnotesize
    \caption{State of the art tools for safety and ethics assessment of Artificial Intelligence (AI)} 
    \begin{center}
    \resizebox{\textwidth}{!}{%
    \begin{tabular}{c|c}
    \hline
    Strategy & Description\\
    \hline 
    \multirow{3}{*}{AI2\cite{gehr2018ai2}} & Scalable analyzer for DNN\\
                       &Proves robustness \\
                       &Uses abstract transformers to capture the behavior of dense and CNN layers\\
                       \hline 
    \multirow{3}{*}{SMLP \cite{brausse2024smlp}} & Symbolic ML prover library\\
                                & Exploration based on data samples \\
                                & Tested in industrial settings at Intel\\
                        \hline
    \multirow{3}{*}{HEFactory \cite{cabrero2023hefactory}} & Symbolic compiler for privacy-preserving DNN\\
                                &Uses homomorphic encryption \\
                                &Obtains 80\% reduction in the number of lines of code\\
    \hline
    \multirow{3}{*}{ConstraintFlow \cite{singh2024constraintflow}} & It is a declarative Domain Specific Language (DSL)\\
                                & Possibility to specify abstract interpretation-based DNN certifiers\\
                                & Lightweight automatic verification of soundness of DNN certifiers\\
                            \hline
    \multirow{3}{*}{CROWN\cite{zhang2018efficient} and Beta-CROWN\cite{wang2021beta} } & Framework for robustness certification/verification of NN\\
                                & Uses bound propagation-based method\\
                                & Flexible on networks with general activation functions (ReLu, tanh, sigmoid and arctan)\\
                               
    \hline

    \multirow{3}{*}{REVISE \cite{wang2022revise}} & Reveling visual biases tool\\
                                &Object-based, person-based, and geometric-based bias detection \\
                                &Suggestion to the user(s) how to mitigate the encountered bias \\
                                \hline
    \multirow{3}{*}{Surprise Adequacy \cite{kim2019guiding}} & Test adequacy criterion for DL systems\\
                                & Different in system's behavior between the input and the training data (Surprise)\\
                                & Systematic samples of inputs based on surprised increased robustness against adversarial samples\\
                               \hline
    \multirow{7}{*}{Ethik AI \cite{bachoc2023explaining}} & Python package "ethik" available \\
                                &Detects model influence concerning protected attributes\\
                                &Identifies causes for why a model performs poorly on certain inputs\\
                                &Visualizes regions of an image that influence a model's predictions\\
                                &Build counterfactual distributions that permit answering "what if?" scenarios\\
                                &Only consider realistic scenarios, and will not build fake examples\\
                                &It scales well to large datasets \\
                               \hline
    \end{tabular}}
    \end{center}
    \label{tab:SafetyEthics}
    \end{table}
\section{AI Assurance (AIA)}
\label{chap:AIA}
\begin{figure}[!t]
    \centering
    \includegraphics[width=\textwidth]{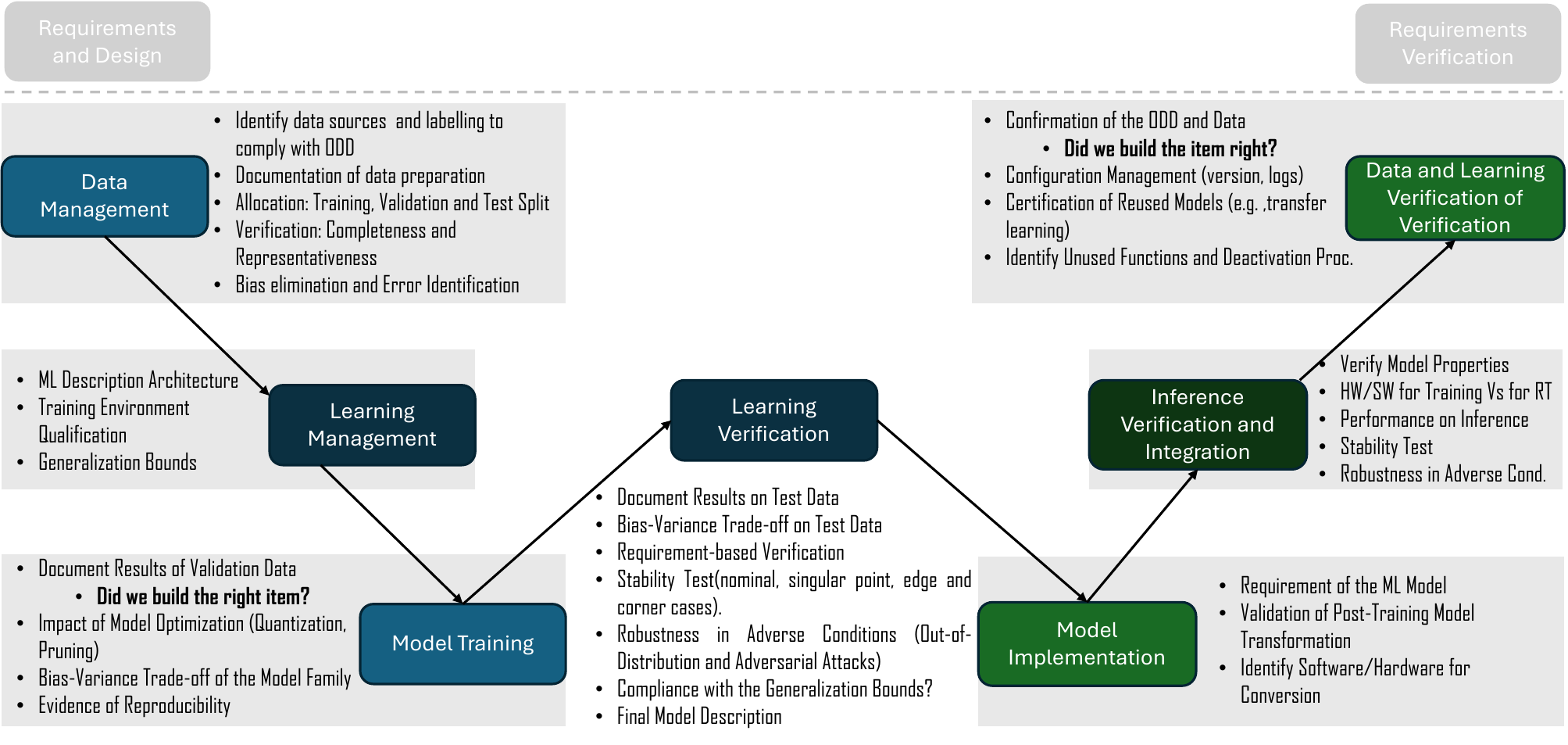}
    \caption{Overview of the learning assurance cycle using the European Union Aviation Safety Agency (EASA) W-shape model. The right side (blue color) shows the offline design of the machine learning (ML) and the left side (green color) represents the cycle towards online deployment.}
    \label{fig:AIAW}
\end{figure}

AI assurance (AIA) defines the objectives of the AI subsystem, employing a system and user-centric approach. 
Two main blocks are identified, Learning Assurance and AI explainability. 
\cref{fig:AIAW} shows the \textbf{W-shape} model for AI assurance of EASA together with an overview of takeaways from each step. 
Below the dotted line are the steps that need to be adapted for AI systems and above it is the traditional assurance cycle. 
As shown in the \cref{fig:AIAW} there is a clear separation between offline (in blue color) and online (in green color) AI assurance process. 
Each step in the cycle is co-dependent and the next step verifies that the previous certification steps are still valid. 

\textit{Data Management} is the first step (under the dotted line) of the \textbf{Learning Assurance} block. 
In this part, the completeness and representativeness of the dataset determine the compliance with the AI constituent's operational design domain (ODD) requirements. 
Completeness indicates that the dataset was reviewed and sufficiently covers the entire space of the defined ODD for the intended application. 
This will ensure the performance of unseen data and help generate generalization bounds for the model. 
Representativeness means that the dataset consists of uniformly distributed and independently sampled data points in the input space, and it is similar to the input space of the intended application. 
The second role of data management is to reduce the impact of the bias. 
Sensors, experiment designs, and data preparation introduce bias to the system. 

For the \textit{Learning Management}, clear and accurate generalization bounds for deep learning models ensure performance on unseen data and help define uncertainty bounds for out-of-distribution data. 
This is particularly important to identify a specific application's singular, edge, and corner cases. 
Furthermore, the \textit{Model Training} step has to be reproducible and report the impact of model optimization techniques, requiring extensive documentation of the validation methods, focusing on answering the question "Did we build the right item?". 
Additionally, to \textit{Verify the Learning} two main tests are introduced: the Stability and Robustness test of the model in adverse conditions, for example, for the case of response against out-of-distribution data and adversarial attacks.

The \textit{Model Implementation} is the first step towards deploying it for online execution. 
In this phase, the AI developer must consider the requirements of the AI/ML model to identify the software/hardware tools needed to convert and use at the time of inference. 
For the \textit{Inference Verification and Integration}, it is necessary to identify the differences between the SW/HW used for training and those used for inference, including the compliance with the performance tolerance defined in the ODD. 
In this step, the stability and the robustness tests are repeated. 

And lastly, it is the \textit{Data and Learning Verification of Verification}. 
Here, comes the answer to the question "Did we build the item right?". 
This is achieved by confirming compliance with ODD requirements and verifying the completeness and representativeness of the data. 
In the case of updating the system, such as reusing the model from another domain (transfer learning), a new certification procedure and a configuration management system are required to record different versions and logs (errors/failures). 
This completes the Learning Assurance block of AIA. 

The second block is the development and post-ops \textbf{AI explainability}. 
This is related to transparency, traceability, safety, security, and accountability of the AI constituent.
It should be noted that the AI system must be interpretable by a wide range of users and personnel from official institutions, such as engineers, certification authorities, and flight crews. 
The wide range of users/stakeholders implies different levels of detail of explainability for each target audience (e.g., specialized EASA personnel or pilots as end users). 
A relevant requirement for the stakeholder is to be able to build trust in the system.
This requires quantifying the confidence level (uncertainty) in the AI system's output.
Uncertainty level and performance need to be continuously monitored during the system's lifetime. 

Overall, AI assurance emphasizes the following objectives: completeness, representativeness,  generalization bounds, stability and robustness of the model, explainability, and continuous monitoring of performance and confidence levels. 
Therefore, research methods with these objectives are presented in the following subsections.

\subsection{Methods for data completeness and representatives}

A trade-off between completeness and representativeness is needed to assure generalization bounds, thus the state-of-the-art approaches of these objectives are merged in \cref{tab:DataComp}. 
This is a requirement from the data management step. 
Real data is high dimensional and faces challenges such as missing values, outliers, noise, and labeling errors. 
The question is how to quantify the trade-off between completeness and representatives in real conditions. 
One way is to reduce the dimensions to visualize the data and discover hidden patterns in the distribution.
A complete and representative dataset has a homogeneous scatter plot. 
There are multiple methods for this and they depend on the data types (text, signal, pixels) and whether linear and nonlinear reduction techniques are necessary \cite{ayesha2020overview, rani2022big}.  

Principal component analysis (PCA) is one of the most used techniques in the literature to find uncorrelated features.
It is a simple visualization technique that removes multicollinearity and reduces parameters and training time. 
The vanilla version of PCA applies to linear datasets and is sensitive to outliers \cite{mackiewicz1993principal}.
At the same time, it is computationally expensive, the new dimensions are not interpretable and there is information loss. 

On the other hand, it is common to assume that the dataset contains independent and identically distributed data points. 
However, this assumption is often violated. 
Sensors monitoring a common phenomenon are interrelated with each other. 
Moreover, a phenomenon in nature involves many interactions between subsystems, e.g., in chemistry molecules will interact with each other in biochemical events. 
The graph-based analysis captures these dependencies. 
It can be used to check the desired coverage of ODD data while filtering out redundant data and enforcing evenly distributed data points. 
In machine learning, exists an entire area dedicated to graph neural networks\cite{zhou2020graph, yun2019graph}. 
\cref{tab:DataComp} mentioned some of the newest techniques in this area. 

A relevant technique is entropy analysis.
This can identify patterns within data by measuring the level of randomness in the dataset. 
It can detect anomalies and group similar data points together, and can then be used to enforce independence between data points. 
The idea will be to homogeneously increase the entropy of the data (e.g., label-wise).
For example, with augmentation techniques, special care must be taken to avoid the addition of outliers, which would result in heterogeneous addition. 
Hence, \cref{tab:DataComp} includes information-aware augmentation techniques. 
These techniques depend highly on the dataset type and require domain expertise\cite{maharana2022review}. 

Another method for high-dimensional data is quantifying the data points' similarity. 
The idea is to reduce and extract meaningful information from the input using latent space (embedding). 
A metric is then used to measure the similarity in the embedding space. 
This method depends on the technique applied to create the latent space and the similarity metric selected\cite{ontanon2020overview,mathisen2020learning}. 
In \cite{geissler2023latent} an interactive latent space an inspector tool is introduced. 
This tool allows AI developers to inspect neural network models' output behavior. 
The user can manipulate values in any latent layer and analyze the response. 
This is a particularly relevant technique to test the robustness of the model against adversarial attacks.  
Exploring the feature space while entering out-of-distribution data can provide information about system behavior at the boundaries, also aiding in fault identification. 
\begin{table}[!t]
    \centering
    \footnotesize
    \caption{State of the art methods for data completeness and representativeness} 
    \begin{center}
    \resizebox{\textwidth}{!}{%
    \begin{tabular}{c|c|c}
    \hline
    Strategy & Work & Main Advantage\\
    \hline 
    \multirow{4}{*}{Principal component analysis (PCA)} &fPCA\cite{palummo2024functional} and O-ALS\cite{gomez2024solving}&Incomplete data-aware\\
                       & UPCA \cite{yao2024unlabeled}& Robust to ground truth corruption\\
                       & OR-TPCA \cite{zhou2017outlier}&Robust to outliers\\
                       & PCA-KPCA\cite{jiang2018parallel} & PCA for linear and nonlinear data\\
                      \hline 
    \multirow{4}{*}{Graph-based} &GNNGuard\cite{zhang2020gnnguard}& Robust to adversarial attacks\\
                                & NetGAN \cite{bojchevski2018netgan}&Add Generalization properties\\
                                & Subgraph Isomorphism counting \cite{bouritsas2022improving}& Add expressivity with symmetry breaking\\
                                & DIGRESS\cite{vignac2022digress}& Robust to noise\\
    \hline
    \multirow{2}{*}{Information-aware augmentation} &Graph-based Entropy aware augmentation \cite{liu2021graph} & Useful for high dimensional data\\
                     &Constrative-based augmentation \cite{luo2023time}  & Time series information-aware\\
                       \hline 
    \multirow{3}{*}{Neuron coverage} & DeepXplore\cite{pei2017deepxplore}& Automated whitebox testing \\
                                    & DeedGauge\cite{ma2018deepgauge}& Multi-granularity testing criteria \\
                                    &DeepTest\cite{tian2018deeptest}, DeepHunter\cite{xie2019deephunter} DeepMutation\cite{ma2018deepmutation}& Detecting erroneous behavior\\

    \hline
    \end{tabular}}
    \end{center}
    \label{tab:DataComp}
    \end{table}
    
Moreover, neural coverage is an attempt to find an intuitive test criterion for a neural network\cite{harel2020neuron}. 
It is based on measuring the proportion of activated neurons (nodes) activated in a forward pass. 
The hypothesis is that a higher proportion implies higher quality. 
It can detect erroneous behavior by generating inputs that maximize the number of activated neurons and then exploring the output layer of the network. 
To measure it, the whole model has to be a white box, and domain expertise is needed to understand its meaning. 

Feature space characterization is a model-centric method capable of determining a dataset's completeness. 
It follows the intuition that a homogeneous feature space indicates a complete dataset (learning-wise), and depends on the task. 
It relies upon metrics such as equivalence partitioning, and pairwise boundary conditioning. 
Equivalence partitioning is a metric to measure the class imbalance, where all labels should converge to one, which is particularly relevant in data clustering \cite{hu2023equivalence}. 
The boundary condition consists of aggregating the limits of each class. 
To do this, the confidence scores between the best and the second guess must be compared\cite{denis2017confidence}. 

The above is a summary of the methods for the objectives of completeness and representativeness of the dataset concerning the ODD, and an overview is in \cref{tab:DataComp}.  

\subsection{Methods for AI generalization, explainability, and uncertainty assurance}
The next objective is the generalization of the AI method. 
Generalization refers to the ability of a model to maintain average performance on unseen data consistently.
A model with generalization properties can handle real-world data variability and will adjust to different operation conditions. 
To evaluate this, the model's training can be explored using learning curves and convergence stability, and then, at test time, the empirical measure of the gap between the training and test data sets can be obtained. 
Some methods exist to increase generalizability. 
These include regularization \cite{santos2022avoiding, kukavcka2017regularization}, stability\cite{zheng2016improving}, deep metric learning \cite{kaya2019deep}, model architecture, and hyperparameter tuning. 
They aim to learn richer network representations to boost performance on unseen data. 
\cref{tab:Generalization} shows an overview of recent methods toward the generalizability of AI models. 
Still, effective methods need to be developed to quantify levels of generalization assurance throughout the learning assurance cycle. 

\begin{table}[!t]
    \centering
    \footnotesize
    \caption{State-of-the-art strategies for data generalization} 
    \begin{center}
    \resizebox{\textwidth}{!}{%
    \begin{tabular}{c|c|c}
    \hline
    Strategy & Work & Main Advantage\\
    \hline 
    \multirow{5}{*}{Regularization} & DropBlock \cite{ghiasi2018dropblock} & Effective dropout for CNN\\
                       & CutMix \cite{yun2019cutmix} &  Combined Mixup\cite{zhang2017mixup} and Cutout\cite{devries2017improved} for augmentation\\
                       & ResNeSt \cite{zhang2022resnest} & Improved diverse representation\\
                       &  RSC \cite{huang2020self}& Improved cross-domain generalization of CNN \\
                       &From Hope to Safety \cite{dreyer2024hope}& Gradient penalization to reduce bias sensitivity\\
                     
                      \hline 
    \multirow{2}{*}{Stability and Robustness} & Squared Residual Network \cite{noorizadegan2024stable} & Enhanced stability in physics-informed neural networks \\
                     & Threshold Networks \cite{ahn2024learning}& Uses the "edge of stability" for generalization\\
                     & Sharpness-Aware Minimization\cite{long2024sharpness} & Gradient-based NN training algorithm to avoid "sharp minima" \\
                       \hline 
    \multirow{3}{*}{Loss function} & Ensemble loss functions \cite{zabihzadeh2024ensemble}& Generalizability-aware for deep metric learning methods \\
                                    & TaskMet \cite{bansal2024taskmet} &  Emphasize learning for the downstream task\\
                                    \hline

    \multirow{3}{*}{Optimizer} &AdaBelief\cite{zhuang2020adabelief} & Fast convergence, generalization and training stability \\
                              &Lion\cite{chen2024symbolic} & Memory-efficient symbolic discovery of optimization algorithms \\
                              &SYMBOL \cite{chen2024symbol}& Automatic discovery of black-box optimizer with symbolic equation learning \\
                                \hline
    
    \multirow{3}{*}{Deep metric learning (DML)} & OBD-SD \cite{zeng2024improving} & Increase embedding diversity \\
                                    & DADA \cite{ren2024towards} & Proxy-based DML to reduce ambiguity \\ 
                                    &Bayesian Metric Learning \cite{warburg2024bayesian} & Uncertainty-aware DML\\
                                    &PRISM \cite{liu2021noise}& Noise resistant technique for DML\\
                                    \hline
    \multirow{3}{*}{Architecture selection} & AutoKeras\cite{jin2023autokeras} & Automated machine learning library \\
                                    & AMLB \cite{gijsbers2024amlb}& AutoML benchmark\\
                                    & Harmonic-NAS \cite{ghebriout2024harmonic} & Hardware-aware multimodal neural architecture search \\
                                    &AZ-NAS\cite{lee2024az}& Training free neural architecture search\\
                        
    \hline
    \multirow{3}{*}{Hyperparameter selection} & PriorBand \cite{mallik2024priorband}& Combined expert belief and proxy tasks\\
                                    & Interactive optimization \cite{giovanelli2024interactive} & Human-centered interactive hyperparameter optimization \\
                                    
    \hline
    \end{tabular}}
    \end{center}
    \label{tab:Generalization}
    \end{table}

Furthermore, AI requires explainability and uncertainty assurance in critical domains such as avionics. 
There are two main types of uncertainty: random and epistemic.
Random uncertainty is known as data uncertainty. 
Epistemic uncertainty implies inadequate knowledge of the AI model \cite{abdar2021review}. 
Explainability refers to interpreting the model's output concisely and user-friendly \cite{adadi2018peeking}. 
Uncertainty meets explainability when accurate prediction and relevant explanations of those predictions are a must. 
This also includes quantifying the uncertainty of explanations and explaining the sources of uncertainty, leading to trustworthy AI. 
\cref{tab:UncertaintyExplainability} comprehends an overview of the latest techniques toward an explainable AI.

\begin{table}[!t]
    \centering
    \footnotesize
    \caption{Uncertainty and explainability methods in the literature} 
    \begin{center}
    \resizebox{\textwidth}{!}{%
    \begin{tabular}{c|c|c}
    \hline
    Strategy & Work & Main Advantage\\
    \hline 
    \multirow{5}{*}{Uncertainty} &NeuralUQ \cite{psaros2023uncertainty,zou2024neuraluq}& Framework for uncertainty quantification\\   
                       &Uncertainty toolbox \cite{chung2021uncertainty}& Open source library for uncertainty quantification (UQ) \\
                       &Beyond Pinball Loss \cite{chung2021beyond} & UQ using full quantile function for regression\\
                       &Transformer Neural Processes \cite{nguyen2022transformer} & Uncertainty-aware meta learning as a sequence modeling problem\\
                       &UR2M \cite{jia2024ur2m}&Resource-aware uncertainty estimation\\         
                       &Uncertainty-aware \cite{sensoy2020uncertainty}& Generative adversarial networks (GAN) for out-of-distribution samples \\
                       &Distance-aware uncertainty\cite{li2024principled}& Enhancing the reliability of physics-informed neural networks(PINNs)\\
                      \hline 
    \multirow{2}{*}{Explainability} & Why Should I Trust You? \cite{ribeiro2016should} & Simple ML model to explain complex DNN outputs\\
                     & Concept based vectors \cite{moayeri2023text2concept,pan2023surrocbm} & Interpretability of the model based on human concepts\\
                     & Concept bottleneck models \cite{shang2024incremental,pan2023surrocbm}& Maps visual representation to human-friendly descriptions\\
                     &StylEX \cite{lang2021explaining}& GAN to explain attributes that underlie classifier decision \\
                     \hline 
    \multirow{3}{*}{Uncertainty meets explainability} & ShapGAP\cite{mariotti2023beyond}& Fidelity measurement to quantify faithfulness of surrogate models \\
                                    & But Are You Sure? \cite{marx2023but} & Uncertainty sets for uncertainty-aware explanations of the models\\
                                    &Explaining the Uncertain \cite{chau2024explaining} & Using Shapley values to explain Gaussian process models\\
                                  \hline
    \end{tabular}}
    \end{center}
    \label{tab:UncertaintyExplainability}
    \end{table}

\section{Human Factors for AI (HFAI)}
\label{chap:HFAI}
\begin{figure}[!t]
     \centering
     \includegraphics[width=\textwidth]{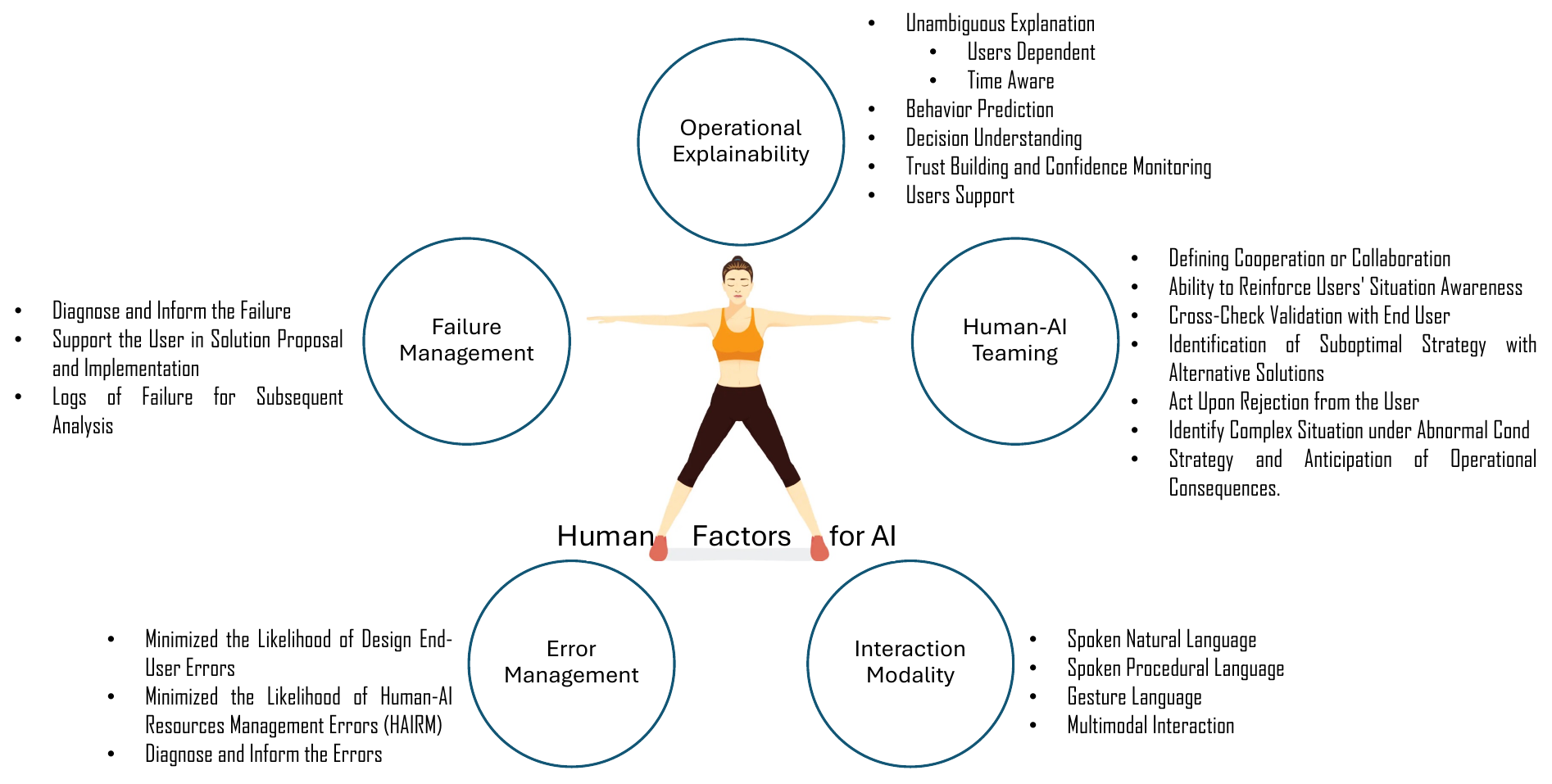}
     \caption{Human factors for artificial intelligence (HFAI) overview}
     \label{fig:HFAI}
 \end{figure}

Human-centered AI focuses on cooperation and collaboration that builds teams of human AI-based systems.
This team encompasses a wide range of end users with diverse skill sets that ally with AI to achieve a goal. 
In the case of cooperation, the AI-based system works as a tool that helps the user(s) fulfill the user's goal. 
In collaboration, the AI and the user(s) work together and jointly to accomplish a shared goal. 
Collaboration implies real-time communication and situational awareness between the AI and the human. 
For the certification cycle to take into account the human factors of AI, five main requirements must be meet: 1. AI operational explainability; 2. Human-AI teaming (collaboration); 3. Modality of interaction and style of interface; 4. Error management; 5. Failure management. 
\cref{fig:HFAI}, highlights the description of the five main requirements. 

The AI system must be equipped with an unambiguous and time-aware explanation of its output to the user, simultaneously with requests for cross-validation by the end user. 
This \textit{operational explainability} depends on the user's level of expertise and the task to achieve. 
The aim is to progressively build trust in the AI system together with confidence level monitoring.
Moreover, a balanced level between the information given to the user and the user's cognitive load is necessary. 
In \textit{human-AI teaming}, the interaction style of the interface varies in terms of modality. 
The modalities include natural language, procedural language, gesture language, and multimodality. 
The selection of one or multiple of those modalities needs to consider the context of the situation to guarantee the performance level under a hostile environment, for example, in the case of a noisy environment and involuntary gestures. 
The AI system has to be able to automatically select the modality(ies) based on the user's state (workload, stress, cognitive resources), situation, and perceived context and adapt to the user's preference. 
In addition, human-AI factors can lead to errors that, undetected, become defects that, in turn, can become failures. 
Consequently is highly relevant to detect, minimize, and provide solution support for errors and failures. 
Human-AI teaming is a system with a huge variety of resources, thus it is a must to employ crew resources management (CRM). 
CRM is the effective use of resources, including people, for a safe and efficient operation. 
This is defined in the SKYbrary website\footnote{https://skybrary.aero/}. 
SKYbrary contains articles related to aviation safety and certification on the topics of operational issues, human performance, enhancing safety, and safety regulations, among others. 
SKYbrary focuses on the usual avionics system, the challenge is how to establish the connection between AI-based avionics and current regulations. 

The human factor in AI has been the subject of numerous studies on how it should be applied, but it remains in question\cite{flathmann2021modeling,zhang2021ideal,hauptman2023adapt}. 
This is because making the same decision without AI is different from making it with AI, and there are still many open issues on how AI works. 
HACO \cite{dubey2020haco} introduces a framework for developing Human-AI teaming using a graphical user interface. 
The authors in \cite{heinzl2024towards} present, a commercial version of an AI platform that provides a solution for human-AI collaboration in manufacturing, called ``Teaming.AI".
They employed knowledge graphs to integrate semantic information of diverse processes executing during runtime. 
In general, human-centric AI will relieve the human decision-maker of pressure. 
However, this could lead to over-reliance on AI predictions, worsening the performance compared to working unassisted. 
In \cite{cabrera2023improving}, the authors aim to improve the decision-making of humans working with AI with the use of "behavior descriptions". 
These descriptions come from the AI developer's mental model, which are details of how the AI performs on subsets of instances \cite{andrews2023role}. 
A trending question is how to use large language models (LLM) to support the Human-AI teaming \cite{vats2024survey}. 
In \cite{mozannar2024effective} the authors use a large language model (LLM) to describe the data regions.
These LLM descriptions are then used to teach the human user through an onboarding stage to improve the human-AI association. 
Moreover, the authors in \cite{ma2024towards} noticed that humans rarely trigger analytical thinking when a disagreement with AI occurs, thus they proposed "Human-AI deliberation" to promote human reflection and discussion related to the AI decision-making process. 
This is aligned with the definition of contestability. 
This means there must be a timely process to allow individuals to challenge the use or outcomes of the AI system. 
Contestable AI is necessary when the AI system significantly affects a person, community, group, or environment \cite{leofante2024contestable}. 
Hence, identifying and addressing users' transparency needs becomes a challenge and a critical element of the Human-AI teaming \cite{turri2024transparency}. 
The goal of these techniques is to help humans recognize when to trust the AI, collaborate/cooperate with it, question it, or ignore and report an AI error. 

\section{AI Safety Risk Mitigation (AIS)}
\label{chap:AIS}
This section addresses the reality of the impracticality that could arise at the moment of certification. 
AI safety risk mitigation is required to counter the fact that exhaustive testing is impossible for complex systems and residual risks remain.
Partially complying with the certification requirements means the entire system has inherent AI risks. 
This is to be expected in black box models such as AI systems. 
Safety risk mitigation is not aimed at compensating partial coverage of objectives belonging to the trustworthiness analysis (TA) certification block i.e., the TA block is critical. 
The purpose is to minimize unexpected/inexplicable behavior of the AI/ML constituent. 
Hence, real-time monitoring and safety net backups (traditional backups) are means to achieve this. 
Still, it is difficult to determine the safety precautions of AI systems due to the newness of AI in the aviation domain and the lack of field experience \cite{MLEAP}. 
\section{Certification Challenges of Classical AI Research}
\label{chap:Example}
\begin{table}[!t]
    \centering
    \footnotesize
    \caption{Trustworthiness analysis overview of classical AI cycle: YOLOv8 Example} 
    \begin{center}
    \resizebox{\textwidth}{!}{%
    \begin{tabular}{c|c|c}
    \hline
    Step & Details/Inquiries & YOLOv8\\
    \hline 
    \multirow{4}{*}{Users}  &List Users and Role & Human and Robot\cite{kumar2024assessing,pitts2024warehouse}\\
                       &Teaming Level & Cooperation\cite{kumar2024assessing,lysakowski2023using}, collaboration\cite{babi2023autonomous}\\
                       &Expertise & --\\
                       &AI Task and User & Object detection. Retail\cite{kumar2024assessing} and warehouse workers\cite{pitts2024warehouse}\\
    \hline
    \multirow{5}{*}{Operational Domain (OD)}  &Input Collection & Camera images and eye tracking\cite{kumar2024assessing}\\
                       &Environment & Retail\cite{kumar2024assessing}, warehouse\cite{pitts2024warehouse}, outdoors\cite{alsamurai2024detection,luo2024novel}, underwater\cite{gupta2024novel,qu2024underwater}\\
                       &Operating Parameters & Light controlled environment \cite{babi2023autonomous,majewski2024end}\\
                       &Degraded Modes & It can be included by retraining \\
                       &OD Completeness and Representativeness& -- \\
                     
                \hline
    \multirow{5}{*}{Safety Assessment }  & Uncertainty Identification & Uncertainty in the detection of edible insects\cite{majewski2024end}\\
                       &Continuous Safety Risk Assessment & --\\
                       &Define Into Service Period&--\\
                       &Out of Distribution Data&--\\
                       &Failure Modes Identification&--\\
                       \hline
    \multirow{7}{*}{Ethics Assessment}  & Interaction with Human & Augmented Reality (AR) and Iris segmentation \cite{lysakowski2023using,lysakowski2023real,zhang2024improved}\\
                       &Robustness and Safety &--\\
                       &Privacy and Data Protection & Privacy-aware YOLOv8 \cite{forster2024decoding,wang2024ai}\\
                       &Transparency & Extensive documentation of YOLOv8 in \cite{yolov8_ultralytics}\\
                       &Non-discrimination & YOLOv8 is trained on large datasets, but bias estimation is overlooked\\
                       &Social and Environmental Well Being & -- \\
                       &Accountability & --\\

\hline
    \end{tabular}}
    \end{center}
    \label{tab:YOLOTA}
    \end{table}

This section presents an example use case to show the certification needs in the classic AI development cycle.
One relevant use case in avionics is collision avoidance systems. 
This includes two steps: 1. detect the object; 2. perform an avoidance maneuver and/or suggest a maneuver to the pilot to satisfy the remain-well clear requirement.
Detecting an object is a task that a human does every day without thinking, making it a comprehensible objective.
Cameras are the most widely used and accurate modality in the literature for solving object detection tasks.

You Only Look Once (YOLO) \cite{redmon2016you} is a vision-based model widely used in object detection, which has multiple versions and has been adapted to embedded implementations \cite{hussain2024yolov1,huang2018yolo,fang2019tinier}. 
YOLO was first published by Joseph Redmon et.al. in 2016, at the time of writing this article, YOLO has already reached version 10. 
Its objective is to predict bounding boxes around objects and class probabilities of the identified object at the same time. 
YOLOv9 \cite{wang2024yolov9} and YOLOv10 \cite{wang2024yolov10} are currently under review, with a reduced number of related publications compared to its predecessor YOLOv8 \cite{yolov8_ultralytics}.
Thus the YOLOv8 is the selected algorithm to analyse.
It focuses on a series of improvements and extensions made by the team of Ultralytics to YOLOv5 \cite{yolov5} and is currently the most stable and widely used version by the research community. 
YOLOv8 series offers object detection, orientation recognition, and object classification, where each variant is optimized for the task.  
Depending on the task, YOLOv8 is trained in a different dataset. 
COCO dataset \cite{lin2014microsoft} or Open Image V7 \cite{OpenImages} for detection, COCO additionally for segmentation and ImageNet \cite{deng2009imagenet} for classification. 
These tasks are relevant in the case of collision avoidance in avionics.

The idea is to identify, based on the analysis carried out within the scope of YOLOv8, the missing certification steps.
The focus is on the detection task to reduce complexity.  
YOLOv8 is a general-purpose object detection algorithm, and many applications/use cases can be defined. 
In addition, the development of YOLOv8 is a community effort, so multiple papers will be cited to show how the authors handle a specific certification step.
Certification queries are performed using synoptic tables to simplify the complex and very dimensional AI verification process. 
In \cref{tab:YOLOTA} is the trustworthiness analysis overview of the selected YOLO version. 
Moreover, in \cref{tab:YOLOWShape} is the W-shape AI assurance queries-based process of the model. 
In \cref{tab:YOLOHuman} is the human-factor-for-AI certification overview of YOLOv8. 
The roadmap for AI safety risk mitigation is not defined at the moment, mainly due to the recentness of AI in aviation and the lack of field experience.

\begin{table}[!t]
    \centering
    \footnotesize
    \caption{AI assurance W-shape overview of classical AI cycle: YOLOv8 Example} 
    \begin{center}
    \resizebox{\textwidth}{!}{%
    \begin{tabular}{c|c|c}
    \hline
    Step & Details/Inquiries & YOLOv8\\
    \hline 
    \multirow{5}{*}{Requirement}  & Safety and Security & -- \\
                       &Functional & Static object detection and moving object detection\cite{safaldin2024improved,yang2024gs}\\
                       &Interfaces & AR\cite{lysakowski2023using,lysakowski2023real} smartphone\cite{shawki2023empowering}\\
                       &Performance metrics & Precision, recall, *mAP, size, parameters, **FLOPS and ***FPS \cite{chen2024yolov8}  \\
                       &Validation & With validation dataset \cite{chen2024yolov8}\\
                     
                      \hline 
    \multirow{5}{*}{Data Management}  & Pre-processing & 640x640 \cite{kumar2024assessing}, 832x832 and frame selection\cite{aboah2023real}\\
                       &Collection & Everyday scenes and natural context\\
                       &Labelling & Manually labeled (validated by visual inspection) \\
                       &Preparation &Feature extraction directly in the model\\
                       &Allocation & Random split 992(train), 124(validation) and 124(test)\cite{chen2024yolov8}\\
                       &Augmentation & Translation, scaling, flipping, mosaic, rotation, cropping \cite{kumar2024assessing,aboah2023real} \\
                       &Completeness and Representativeness test & --\\
                       &Bias Elimination & Trained on large datasets, but bias estimation is overlooked\\
                   &Confguration Management (CM)& Multiple datasets to train YOLO$\bigstar$, but CM is unclear/lacking.\\
                       
    \hline
    \multirow{4}{*}{Learning Management}  & Model Family & YOLO version 8 \\
                       &Learning Algorithm & Stochastic gradient descent (SGD) \cite{chen2024yolov8} \\
                       &Optimizer & Adam; $\dagger$lr 0.0106, momentum 0.971, weight decay 0.00048 \cite{aboah2023real}\\
                       &Parameters Initialization & Warmup epochs 2.689 and $\ddagger$IoU 0.912 \cite{aboah2023real}\\
                       &Generalization Bounds Identification & --\\
                \hline
    \multirow{3}{*}{Model Training} & Cost/Loss Function Curve & Available in Ultralytics training process \cite{yolov8_ultralytics}\\
                       &Optimization Technique & 8-bit fix-point data per-group quantization \cite{wang2024energy}\\
                       &Comparison Between Model Family & Yes; time, parameters, and complexity\\
                       &Reproducible & Yes, with a research community working on it\\
    \hline
        \multirow{3}{*}{Learning Verification}  & Test Results & mAP = 0.5861 @ 95fps\cite{aboah2023real}\\
                         &Robustness in Adversity? & --\\
                         &Compliance with Generalization Bounds? & --\\

                       \hline
        \multirow{6}{*}{Learning Verification (Stability) } 
                       &Identification of Edge/Corner Cases? & --\\
                       &Data Point Replacement? & --\\
                       &Additive Noise Effect? & -- \\
                       &Labelling Errors Induced?& -- \\
                       &Random Initialization Avoided? & Option to enable/disable it\\
                       &Hyperparameter Tunning Stable? & -- \\
            \hline
        \multirow{7}{*}{Model Implementation}  & HW Performance & ***FPS = 67.1 for object detection\cite{wang2024energy}\\
                       &Conversion Method & LLVM-C2RTL toolkit  \\
                       &Optimization for HW & NN layer optimization and $\star$PLF\cite{wang2024energy} \\
                       &Processing Power & RISC-V\cite{wang2024energy} \\
                       &Parallelization & GPU Nvidia \cite{elhanashi2024telestroke}\\
                       &Latency & ***FPS = 67.1 for object detection \cite{wang2024energy}\\
                       &Worst Case Execution Time &--\\
                       \hline
        \multirow{3}{*}{Inference Verification}  & Stability Test? & -- \\
                       &Robustness in Adverse Conditions? & --   \\
                       &Performance on Inference & ***FPS = 67.1 \cite{wang2024energy}, **FLOPS = 8.7 \cite{elhanashi2024telestroke}\\
                       \hline
        \multirow{2}{*}{Verification of Verification} 
                       &Robustness in Adverse Conditions? &--\\
                       &Identify Unused Function? & -- \\
\hline
    \end{tabular}}
    {\centering \par *mAP: Mean average precision. **FLOPS: Floating-point operations per second. ***FPS: Frame per second
    $\bigstar$\url{https://docs.ultralytics.com/datasets/}$\dagger$ lr: Learning rate. $\ddagger$IoU: Intersection over union. $\star$PLF: Piecewise linear function approximation. \par}
    \end{center}
    \label{tab:YOLOWShape}
    \end{table}

To populate some of the entries in the certification tables it has been necessary to include multiple research projects based on YOLOv8. 
A singular research project based on YOLOv8 has no more than five subcells in the tables. 
This reveals the common research practice of focusing entirely on performance metrics, neglecting the assessment of trust and ethics.
In the trustworthiness analysis of YOLOv8 (see \cref{tab:YOLOTA}) the research community assumed that the completeness and representativeness of the dataset are certain by fitting the model for a specific task. 
Pre-training the model using a massive dataset is considered a sufficient method to achieve a complete and robust solution.
Moreover, the safety risk assessment of the deep learning model is commonly overlooked. 
This includes critical aspects such as the uncertainty of model results and the identification of failure modes.
Ethical evaluation is negatively affected in the current AI research cycle, although AI is spreading pervasively into everyday tasks, but remains unaccountable for how it affects humans. 

\cref{tab:YOLOWShape} is evident that most attention is on data pre-processing and techniques to increase performance, for example, accuracy, precision, recall, and F1-score. 
Identifying adverse design responses and compliance with learning verification goes unnoticed.
This leads to unstable and unpredictable AI design, which makes the whole AI modeling effort impractical and risky to cross the research frontier and become a practical application.
Moreover, \cref{tab:YOLOHuman} shows that the certification of human factors for AI in the YOLOv8 case is scarce and needs urgent attention in the AI development process, where error/fault identification and management are crucial to ensure safety in critical applications. 
YOLO is one of the most widely used algorithms in the research community for object detection and classification and has been developed by a large number of researchers over the years. 
Hence, YOLO as a certification example is meaningful. 
The intention is to present an overview of the missing steps and to raise awareness of the need for certification throughout the AI development cycle, where performance metrics are no longer sufficient to conclude a research project. 
Although the use case in this paper is avionics applications, this analysis is fundamental for AI to be used in any critical domain, such as automotive, communication, medicine, and human well-being.

\begin{table}[!t]
    \centering
    \footnotesize
    \caption{Human Factor for AI overview of classical AI cycle: YOLOv8 Example} 
    \begin{center}
    \resizebox{\textwidth}{!}{%
    \begin{tabular}{c|c|c}
    \hline
    Step & Details/Inquiries &YOLOv8\\
    \hline 
    \multirow{4}{*}{Operational Explainability}  & Unambiguous Explanation & --\\
                       &Behaviour Prediction & --\\
                       &Decision Understanding &--\\
                       &Trust Building & --\\
                       &Confidence Monitoring & Confidence based on Intersection over union (IoU)\\
                       &User Support & --\\
    \hline
    \multirow{4}{*}{Human-AI Teaming}  & Cooperation/Collaboration & Depends on the application\\
                       &Reinforce User Situation Awareness & -- \\
                       &Cross-Check with the User & --\\
                       &Identification of Sub-optimal Strategy & -- \\
                       &Act Upon Rejection from the User & --\\
                       &Identify Complex Situations & -- \\
                       &Anticipation of Operational Consequences & -- \\
    \hline
    \multirow{4}{*}{Interaction Modality}  & Spoken Natural Language & -- \\
                       &Spoken Procedural Language & -- \\
                       &Gesture Language & --\\
                       &Multimodal Language & Visual; object image with confidence level \\
    \hline
    \multirow{4}{*}{Error Management}  & Minimized Likelihood of User Errors  &--\\
                       &Minimized Likelihood of Resource Management Errors &--\\
                       &Diagnose and Inform Errors &--\\
                     
    \hline
    \multirow{4}{*}{Failure Management}  & Diagnose and Inform Failures& --\\
                       &Support User in Solution Proposal and Implementation& -- \\
                       &Logs of Failure for Analysis& --\\

\hline
    \end{tabular}}
    \end{center}
    \label{tab:YOLOHuman}
    \end{table}

\section{Discussion and insights}
\label{chap:Discussion}
Avionic is the leading safety-critical domain in AI\cite{EASAConcept}. 
In addition, aviation is one of the most regulated areas for development, with multiple public agencies and users involved in the process. 
Despite the above, the status of AI in avionics is in its infancy. 
The structure of the minimum requirements for certification is currently being outlined. 
The complexity of critical sectors and the lack of AI certification make AI-avionics teamwork extremely delicate. 
Therefore, collaboration between industry, government, and researchers is crucial to identify effective and feasible means of meeting the defined certification objectives. 
This section presents a summary of the limitations of the certification of AI in aviation. 
Due to the sheer size and complexity of avionics and AI systems, this \textbf{list of insights} is far from complete, but it offers a glimpse of what to expect on the road toward certifiable AI. 

\begin{itemize}
    \item \textbf{Generalization of methods:} The certification process and sub-processes are not generalizable. 
    It is a high-dimensional problem that needs tailored assessment methods for application and domain, demanding intensive time-consuming efforts.  
    This disrupts the classic cycle of research advances, in which the most cited projects are general-purpose models. 
    The general purpose modeling style requires a huge effort for certification due to the common practice of bypassing certification in the development cycle and mistakenly assuming that the design only has to meet the output performance metrics. 
    This leaves elements such as ethics, and safety and risk assessment unattended. 
    Certification should be considered from the beginning of AI development. 
    Novel algorithms are constantly being released without being accountable to any of the certification blocks.

    \item \textbf{Operational design domain description (ODD):} The lack of OD and ODD description in the DNN development process greatly affects the completeness and representativeness of the dataset selection.
    Furthermore, the type of data also influences the model structure and parameter settings.
    Consequently, the whole process risks becoming worthless or meaningless, because in the end it does not solve a practical application in a certifiable way. 
    Moreover, without a correct OD and ODD, it is impossible to identify singular point, edge, and corner cases to test the robustness and stability of the system.
    \item \textbf{New learning paradigms:} 
    A variety of deep learning methods are proposed at an incredibly fast pace. 
    Particularly, there is growing interest in new ways to improve the learning capabilities of the model. 
    Due to the large DNN community, it is challenging to list all new methods. 
    Therefore, to reduce complexity the focus will be on four areas of interest: 1. guidance/teaching models; 2. contrastive models; 3. expert knowledge models; 4. autonomous learning. 
    
    Among \textit{teacher models}, transfer learning (TL), and knowledge distillation (KD) exist. 
    DNN models require a large amount of data to converge, hence multiple methods are proposed to mitigate the requirement of large datasets for each specific task.
    The TL process requires two steps: the first step consists of selecting or training a network in a domain where a large dataset is available. 
    The second step consists of fine-tuning the last layers (re-training) of a pre-trained neural network (old domain) using data from the new domain/task\cite{zhuang2020comprehensive}. 
    This method requires a certification procedure for the new domain/task despite being certified in the old/mother domain/task. 
    KD offers the perks of transferring knowledge from a cumbersome model (teacher) to a smaller and more manageable neural network model (student). 
    In this way, the student can learn faster with the teacher's regularization, and the computational complexity and size are reduced, which at the same time can increase the interpretability of the solution. 
    This property is important at the time of model deployment on constrained hardware devices\cite{gou2021knowledge}. 

    \textit{Contrastive learning(CL)} is a deep learning methodology where the network learns by comparison among different input samples. 
    The comparison can be between similar/dissimilar pairs of data points. 
    With this method, the NN learns to push together similar samples and pull away the dissimilar points. 
    For an efficient learning process the selection of the positive/negative samples is crucial. 
    This depends on designing the similarity distribution so that positive pairs are different in the input space but are still semantically related, and on a dissimilarity distribution that ensures that negative pairs are similar in the input space but are semantically unrelated\cite{le2020contrastive}.
    In \cite{winkens2020contrastive,williams2021fool} the authors use CL for out-of-distribution data detection, and in \cite{he2023clur} uncertainty estimation is assisted by contrastive learning. 
    Therefore, CL can be used in the analysis of the completeness and representativeness of the dataset. 
    
    Despite their advantages, the above methods are of great complexity and are mostly conceived without expert knowledge to add explanatory power. 
    On the other hand, researchers are joining efforts to build models with some explanatory meaning based on \textit{expert knowledge} from other disciplines.
    Spiking neural networks (SNNs) are an example of extending the power of NNs by replicating brain behavior as an organic network. 
    This coincides with the main goal of NNs, which are supposed to mimic neural connections in the brain, including interaction and reaction between them.  
    SNNs exist since spikes of biological neurons are sparse in time and space, and event-driven, which is closer to how the human brain computes at the neural description level. 
    SNNs employ bio-plausible local learning rules, making them suitable to build low-power neuromorphic hardware for SNNs\cite{tavanaei2019deep}. 
    Biologically plausible local learning rules can increase the robustness of NN to noise without sacrificing the performance of the task, as synaptic balancing\cite{stock2022synaptic}. 
    Evolutionary algorithms (EA) are also an example of methods based on the principle of biological evolution.  
    EAs can be used as a computational optimization to improve the population of potential solutions iteratively, making them suitable for improving hyperparameters with an objective function\cite{song2024reinforcement,pratap2024optimizing}. 
    Physic-informed neural networks (PINNs) encode physics laws in the form of partial differential equations, which are then used as an additional loss term in the loss function when training the neural network. 
    The learning capability of deep neural networks depends on the size of the dataset. 
    PINNs help to converge the model with a small number of samples without violating known physical laws (added as terms in the loss function)\cite{li2024principled,donnelly2024physics}. 
    Expert knowledge can be represented as rule-based systems, which is the case of symbolic artificial intelligence (SAI).
    It offers a set of methods based on high-level symbolic representations of problems, logic, and search. 
    SAI copes with the unsustainable computational resources of DNN development while adding properties of robustness and explainability to the AI cycle. 
    The combination of NN and symbolic approaches can impact human-AI collaboration with reasoning and cognitive capabilities within AI development\cite{wan2024towards,wu2024symbol,dinu2024symbolicai,cambria2024senticnet}.

    The fourth area is autonomous learning. 
    These are methods that enable AI to learn tasks autonomously. 
    Reinforcement learning (RL) is a powerful method to fully automate AI models. 
    An interesting sub-field of RL is explainable reinforcement learning (XRL). 
    This area aims to understand the decision-making process of RL agents, adding interpretability to these methods helps the use of them in critical domains. 
    In \cite{milani2024explainable,glanois2024survey} the authors present a survey of the techniques, challenges, and opportunities of XRL. 
    In \cite{panoutsakopoulos2022towards} a team of researchers present an RL application for an autonomous Airborne Collision Avoidance System. 
    They use expert knowledge for their model by defining airspace characteristics and aircraft models. 
    They employ a summary of basic concepts of relative geometry and kinematics, adding reliability to the system. 
    In addition, continuous reinforcement learning offers the idea of never stopping learning new tasks, in contrast to typical RL, which consists of finding/improving solutions on predefined tasks\cite{abel2024definition}. 
    In general, despite the advantages of autonomous AI, it also involves additional unknown certification steps. 
    This area is within the next round of discussion by aviation regulators.

    \item \textbf{Explainability:} Deep neural networks are astonishingly increasing in size and complexity while understanding why the new method performs best remains a mystery. 
    This is connected to the need for contestable AI systems.
    Contestable AI becomes more important when an AI system significantly affects an individual, community, group, or environment. 
    In this context, a timely process must allow individuals to challenge the use or results of the AI system. 
    This requires a dynamic relationship between human and AI methods to explain/revise their decision-making process\cite{leofante2024contestable} progressively. 
    
    \item \textbf{AI system definition:} The definition of the AI system and subsystems varies according to the specific avionics domain.
    It could include the AI-human interaction, requiring human-AI teaming accountability. 
    Moreover, AI development needs to quantify the emotional intelligence requirement to understand and manage the human-AI interaction.
    \item \textbf{Automated Machine Learning (AutoML):}
    AutoML is used to generate and optimize AI models. 
    It includes parameter selection/optimization, and an automatic neural architecture search (NAS). 
    A successful AutoML tool should reinforce the researcher's trust, making clear the need for transparency in the development process\cite{drozdal2020trust,zoller2022xautoml}.
    This leads to inquiries such as: Can AutoML be relied upon to speed up the certification process of requirements definition and compliance? Is it possible to include the description of the users and the operational domain in the cycle? Is it possible to automatically select the AI classification? Can Fairness be automated with the use of AutoML?\cite{weerts2024can,amirian2021two}
    \item \textbf{Environmental and well-being:}
    The research community focuses primarily on performance. 
    Currently, performance improvement translates into the use of massive models, which also require enormous use of resources.
    This urgently claims for techniques that advance in AI in an environmentally responsible manner\cite{wu2022sustainable,van2021sustainable}. 
    Network training is oblivious to the resources and energy consumption requirements.  
    Training includes designing huge models by trial and error and tuning hyperparameters, which consumes a large amount of energy\cite{heydarigorji2020hypertune,geissler2024power}. 
    \item \textbf{Failure/error detection and management:}
    To ensure safe operations, the DNN and the system must undergo rigorous verification and validation, including advanced statistical analysis. 
    The performance and safety of the DNN and the system's behavior must be analyzed for the nominal case and in numerous outlier and failure cases. 
    This is part of the safety assurance of the AI system. 
    The definition of safety by researchers mainly refers to the use of the DNN model for safety tasks, without assessing the compliance of the DNN method with safety standards. 
    \item \textbf{Unbalanced attention on certification blocks:} 
    The AI assurance block receives the most attention from the research community. 
    This is mainly due to the close relationship between the AI assurance block and performance improvement. 
    The performance improvement of a model compared to related work is currently the main metric to be accepted by the research community. 
    Meanwhile, ethical and human factors, such as emotional intelligence and training requirements, and managing the security risks of AI solutions are underrepresented and urgently need attention by the community.     
\end{itemize}

\section{Conclusion}
\label{chap:Conclusion}
Avionics is one of the leading critical domains in artificial intelligence (AI), yet its integration is in its early stages.
Aviation, one of the most regulated sectors, involves numerous public entities, making AI certification particularly challenging.
The framework for AI certification in avionics is still being developed, and the complexity of the field, combined with the absence of established AI certification, demands careful collaboration between industry, government, and researchers.
This work outlines the current state of AI certification in avionics, summarizes key certifiable AI components, and highlights the importance of a clear development roadmap.
The findings underscore that AI certification is essential not only for avionics, but for any safety-critical domain such as automotive, communication, medicine, and human welfare.



\bibliographystyle{ACM-Reference-Format}
\bibliography{sample-base}


\end{document}